\documentclass{ieeeaccess}
\usepackage{cite}
\usepackage{amsmath,amssymb,amsfonts}
\usepackage{algorithmic}
\usepackage{graphicx}
\usepackage{textcomp}
\usepackage{adjustbox}
\usepackage{url}
\usepackage{bm}
\usepackage{multirow} 

\usepackage{hyperref}

\makeatletter
\AtBeginDocument{\DeclareMathVersion{bold}
\SetSymbolFont{operators}{bold}{T1}{times}{b}{n}
\SetSymbolFont{NewLetters}{bold}{T1}{times}{b}{it}
\SetMathAlphabet{\mathrm}{bold}{T1}{times}{b}{n}
\SetMathAlphabet{\mathit}{bold}{T1}{times}{b}{it}
\SetMathAlphabet{\mathbf}{bold}{T1}{times}{b}{n}
\SetMathAlphabet{\mathtt}{bold}{OT1}{pcr}{b}{n}
\SetSymbolFont{symbols}{bold}{OMS}{cmsy}{b}{n}
\renewcommand\boldmath{\@nomath\boldmath\mathversion{bold}}}
\newcommand{\orcidicon}[1]{\href{https://orcid.org/#1}{\includegraphics[width=8pt]{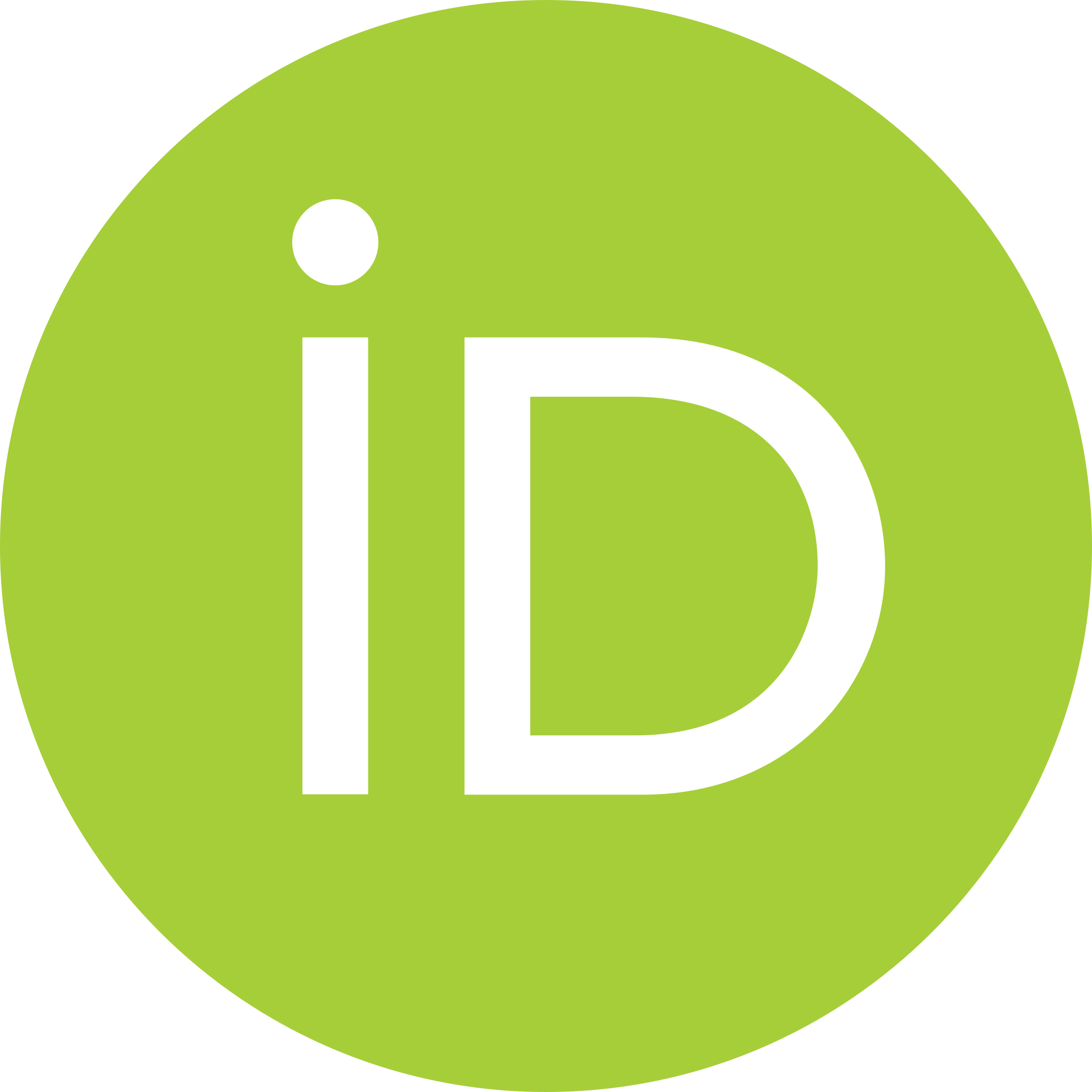}}}

\makeatother

\def\BibTeX{{\rm B\kern-.05em{\sc i\kern-.025em b}\kern-.08em
    T\kern-.1667em\lower.7ex\hbox{E}\kern-.125emX}}

\begin{document}
\history{Date of publication xxxx 00, 0000, date of current version xxxx 00, 0000.}
\doi{10.1109/ACCESS.2024.0429000}

\title{Multi-Task Faces (MTF) Data Set: A Legally and Ethically Compliant Collection of Face Images for Various Classification Tasks}
\author{\uppercase{Rami Haffar}\orcidicon{0000-0002-0139-246X}, 
\uppercase{David Sánchez}\orcidicon{0000-0001-7275-7887},\IEEEmembership{Senior Member,~IEEE} and \uppercase{Josep Domingo-Ferrer}\orcidicon{0000-0001-7213-4962},\IEEEmembership{Fellow,~IEEE}}

\address{Universitat Rovira i Virgili, Department of Computer Engineering and Mathematics, CYBERCAT-Center for Cybersecurity Research of Catalonia, Av. Pa\"{\i}sos Catalans 26, 43007, Tarragona, Catalonia (e-mails: \{rami.haffar, david.sanchez, josep.domingo\}@urv.cat)}

\tfootnote{This research was funded by the the European Commission (project H2020-871042 ``SoBigData++''), the Government of Catalonia (ICREA Acad\`emia Prizes to J. Domingo-Ferrer and to D. S\'anchez), MCIN/AEI/ 10.13039/501100011033 and ``ERDF A way of making Europe'' under grants PID2021-123637NB-I00 ``CURLING'' and {PRE2019-089210}, and INCIBE and European Union NextGenerationEU/PRTR (project ``HERMES'' and INCIBE-URV Cybersecurity Chair).}

\corresp{Corresponding author: Rami Haffar (e-mail: rami.haffar@urv.cat).}

\begin{abstract}
Human facial data offers valuable potential for tackling classification problems, including face recognition, age estimation, gender identification, emotion analysis, and race classification. However, recent privacy regulations, particularly the EU General Data Protection Regulation, have restricted the collection and usage of human images in research. As a result, several previously published face data sets have been removed from the internet due to inadequate data collection methods and privacy concerns.
While synthetic data sets have been suggested as an alternative, they fall short of accurately representing the real data distribution. Additionally, most existing data sets are labeled for just a single task, which limits their versatility. To address these limitations, we introduce the Multi-Task Face (MTF) data set, designed for various tasks including face recognition and classification by race, gender, and age, as well as for aiding in training generative networks.
The MTF data set comes in two versions: a non-curated set containing 132,816 images of 640 individuals, and a manually curated set with 5,246 images of 240 individuals, meticulously selected to maximize their classification quality. Both data sets were ethically sourced, using publicly available celebrity images in full compliance with copyright regulations.
Along with providing detailed descriptions of data collection and processing, we evaluated the effectiveness of the MTF data set in training five deep learning models across the aforementioned classification tasks,
achieving up to 98.88\% accuracy for gender classification, 95.77\% for race classification, 97.60\% for age classification, and 79.87\% for face recognition with the ConvNeXT model. Both MTF data sets can be accessed through the following link. \url{https://github.com/RamiHaf/MTF_data_set}
\end{abstract}


\titlepgskip=-21pt

\maketitle

\section{Introduction}
\label{sec: intro}
\PARstart{A}{rtificial} intelligence (AI) is highly dependent on the availability of data, which is the cornerstone for training and evaluating AI models and unlocking their full potential \cite{jain2020overview}. Since the accuracy and the effectiveness of AI models are directly influenced by the quality of the data on which those models are trained, meticulous collection and precise labeling of data are paramount in machine learning.

Data collection procedures vary significantly depending on the type of data gathered. In the case of tabular data, collecting information can be done explicitly through surveys and polls that require people to answer. An alternative procedure is to obtain this type of data from public (e.g., government) administrative records. When it comes to textual data, there are significant limitations due to copyright restrictions. The right to copy, digitize, and collect text is heavily curtailed by copyright holders \cite{ducato2019limitations}. Nevertheless, there are valuable sources of text data that are not copyrighted or available as public domain material. For example, Wikipedia provides enormous and curated amounts of text data that can be used for training purposes. The situation is similar for image data, where there are restrictions related to the rights to copy and modify images. Yet, one difference between text and image data is that public-domain image sources are much more limited.

Images are especially complex data. On the one hand, their redistribution may or may not be authorized according to the applicable copyright licenses \cite{voyatzis1999protecting}. On 
the other hand, they may be subject to special regulations according to their contents. For instance, medical images are essential to train computer-aided diagnosis systems, but their use requires either explicit consent by the patients or the removal of image traits or metadata that might lead to privacy disclosure \cite{philbrick2019ril}. Also, images containing human faces are employed in a wide
range of applications ---including automated identification of people--- that raise profound legal and ethical issues. In particular, their use is restricted by regulations concerning the protection of personally identifiable information, such as the General Data Protection Regulation (GDPR) \cite{voigt2017eu} or the Ethics Guidelines for Trustworthy AI \cite{smuha2019eu} of the European Union.

More specifically, the GDPR defines personal data in its Article 4
as ``any information relating to an identified or identifiable natural person (`data subject').'' Based on this definition, we can confidently state that facial images are legally considered personal data because they have the capability to easily identify the individuals they depict. The GDPR also limits the processing of personal data to ensure privacy. In Article 9, it states that ``processing of personal data revealing racial or ethnic origin, political opinions, religious or philosophical beliefs, or trade union membership, and the processing of genetic data, biometric data for the purpose of uniquely identifying a natural person, data concerning health or data concerning a natural person’s sex life or sexual orientation shall be prohibited.''
Facial images, due to their re-identifying nature and the information they contain about an individual's ethnic origin, race, and biometric features fall into this category of non-processable data. However, the GDPR indicates exceptions to this prohibition in the same article, specifically in its second paragraph. One such exception is ``(e) processing relates to personal data which are manifestly made public by the data subject.''
Hence, processing of personal data on individuals who have chosen to make them public without any restrictions is allowed. This applies to images that have been published under public-domain licenses or with Creative Commons licenses that grant the right to share and modify the content.

Face images are crucial in many areas of AI research, such as facial recognition \cite{boutros2023synthetic}, emotion detection \cite{chowdary2023deep}, age estimation \cite{gupta2023single}, gender classification \cite{hassan2024face}, facial biometric analysis \cite{sanil20232d}, and other related tasks, such as face anonymization \cite{barattin2023attribute}. 
However, the availability of such data has significantly decreased in the last years.

In particular, several data sets of face images have recently become access-restricted or have been completely removed from the internet due to privacy concerns or to comply with data protection regulations. Furthermore, many image data sets lack clarity regarding the methods employed for image collection, which makes them susceptible to potential removal in the near future.
The withdrawal of previously public data sets has several shortcomings and, in particular, hampers the reproducibility of research results obtained from those data. As an alternative, some data sets are composed only of synthetic images, whose applicability is limited because they tend to deviate from real-world data distributions. 

\subsection*{Contributions and plan}
To overcome the limited availability of human face data sets, in this paper, we present the Multi-Tasks Faces (MTF) data set, a collection of real-face images aimed at training and evaluating AI models for various tasks, including face recognition (FR), gender classification (GC), age classification (AC), and race classification (RC). The data set is presented in two flavors: a non-curated version that includes 132,816 images of 640 individuals, and a manually curated version with 5,246 images of 240 individuals meticulously selected to maximize their classification quality.
The compilation of the images has been done in compliance with current legal regulations. 
In particular, all original images are publicly available and correspond to well-known celebrities, which avoids identity disclosure issues while keeping data identified. 
In addition, the compiled images have a copyright license that is either public domain or Creative Commons with permission to modify, share, and commercial use.  
Both data sets have been compiled under the umbrella of the SoBigData++ project \footnote{\url{https://plusplus.sobigdata.eu/}}, and they have been approved by the ethical and legal board of the project. SoBigData++ is a project funded by the European Union that aims to design and deploy an integrated infrastructure for social mining and big data analytics, and it adheres to the world's strictest data protection regulations.

In addition to describing the data collection and processing procedures that were carried out to build the data sets, we also report the results
of a comprehensive evaluation of the curated and non-curated MTF data sets on several well-known deep learning (DL) models. These results constitute baselines for future research endeavors that employ images for the intended tasks. Furthermore, we highlight the significance of manual data processing by comparing the performance of each DL model trained on the curated MTF data set with the same model trained on the non-curated version. This justifies the necessity and importance of the different steps followed to create the curated MTF data set.

The remainder of this paper is organized as follows. Section \ref{sec: Related} discusses previously published data sets of face images and highlights their limitations. Section \ref{sec: methodologyu} details how our data sets have been collected, processed, and labeled. Section \ref{sec: data Specification} describes the structure of the data sets in detail. Section \ref{sec: results} reports the baseline results obtained when using the curated MTF data set with five different DL models, and compares its results with the non-curated version. The final section summarizes the key contributions of this work and outlines several lines for further research.

\section{Related work}
\label{sec: Related}

CMU Face Images \cite{CMUface} is a small data set consisting of 640 black and white face images corresponding to 20 individuals (32 images per individual). It is labeled according to pose (straight, left, right, up), expression (neutral, happy, sad, angry), eyes (open, sunglasses), and scale (full-resolution, half-resolution, quarter-resolution). Information is not available on the type of consent the 20 individuals gave in regard to releasing their images. On the other hand, this data set allows tasks related to pose/emotion
classification, but its small size, its lack of race diversity, the low resolution of some images, and the fact that they are black and white limit its applicability. 

The Large-scale CelebFaces Attributes (CelebA) data set
\cite{liu2018large} is a widely employed data set consisting of 202,599 face images featuring 10,177 celebrities. Each image is labeled according to the corresponding identity, in addition to 40 binary attributes depicting the appearance of the individual (e.g., eyeglasses, bangs, wearing hats, mustaches), among others. However, the CelebA data set was compiled by crawling images from the Internet, irrespective of the copyright ownership held by the owners of the images. Consequently, the availability of the CelebA data set is potentially at risk due to privacy concerns and copyright laws, particularly within the European Union \cite{rosati2019copyright}.

The recently published DigiFace-1M data set \cite{bae2023digiface1m} comprises 1.22 million synthetic face images. It is divided into two parts: i) 720,000 images featuring 10,000 virtual identities (72 images per identity), and ii) 500,000 images encompassing 100,000 virtual identities (5 images per identity). Although this data set offers a substantial number of images, it lacks information regarding the distribution ratio between males and females or between different age groups. It is important to note that the sole purpose of this data set is to train FR models. However, it is worth considering that all the images in this data set are synthetically generated. This fact limits the data set's representativeness, as it does not accurately reflect real-world data.

The Labeled Faces in the Wild (LFW) data set \cite{LFWTech} is another noteworthy data set of face images. It contains 13,233 images, corresponding to 5,749 unique identities. The publishers of the data set emphasize that all the data was legally and ethically collected from original images under the Creative Commons copyright licenses. However, LFW presents two significant drawbacks. Firstly, the images are not `cleaned', meaning that the backgrounds of the images contain a substantial amount of information, sometimes including other individuals' faces. Secondly, each identity in the data set has a very limited number of images, with only 1,680 identities having more than one image and none of the identities having more than three images. These drawbacks limit the applicability of this data set to certain unsupervised biometric identification applications and render it unsuitable for training supervised models.

Another accessible data set is Megaface \cite{kemelmacher2016megaface}, which consists of one million face images that have been ethically collected from images that fall under Creative Commons copyright licenses. However, Megaface is specifically designed to facilitate the training of \emph{face detection} models. The data set exclusively offers bounding boxes that specify the facial regions within the images, while it does not include identity labels for face recognition purposes. This makes Megaface a valuable resource for advancing face detection algorithms, but it may not be suitable for tasks that require identity recognition or other demographic classification tasks.

Similarly to Megaface, other data sets have been released with the specific purpose of facilitating the training of face detection models. Two such data sets are the WIDER FACE data set \cite{yang2016wider}, which comprises 32,203 images, and the Multi-Attribute
Labeled Faces (MALF) data set \cite{faceevaluation15}, which comprises 5,250 images.
Both of these data sets have been carefully collected, ensuring ethical collection practices, and each image has been labeled to indicate the precise location of faces within the images. However, both data sets are exclusively suitable for face detection tasks and cannot be directly employed on any other AI training tasks without additional processing and relabeling of the images.

IMDB-WIKI \cite{Rothe-IJCV-2018} consists of 523,051 face images, which were obtained from two well-known websites (IMDB and Wikipedia) providing images from the public domain. Nevertheless, this data set is only labeled for age estimation.

The Flickr-Faces-HQ (FFHQ) data set \cite{karras2019style} is an extensive collection of 70,000 high-quality images portraying human faces with remarkable diversity in terms of age, ethnicity, and image backgrounds. These images were directly downloaded from Flickr,
 and they are subject to Creative Common copyright licenses. Additionally, FFHQ lacks any processing or labeling, as its primary purpose is to aid in the development of generative adversarial networks.

On the other hand, several other popular face image data sets have been confronted with accessibility challenges due to the way their images were compiled.
VGGFace2 \cite{cao2018vggface2} featured 3.3 million face images captured in real world conditions, representing more than 9,000 distinct identities. This data set was specifically labeled for FR, with an average of 362 images per identity. However, the images within the data set were obtained by downloading them from Google image search without taking into account the copyright ownership of the images. As a result, VGGFace2 is no longer publicly available. Although the specific reasons for its withdrawal from public access have not been explicitly disclosed, they are likely to be related to concerns about privacy and legal considerations.

UMDFaces \cite{bansal2017umdfaces} is another significant data set that contains 367,888 facial images of 8,277 individuals, which is labeled for face recognition, but also provides information on 21 key points that capture essential biometric details. However, the authors have not specified the methodology employed to collect and process the data, thereby raising concerns about the ethical aspects of the creation of this data set.
Currently, UMDFaces is not available for download. The data set's website indicates that the authors are working towards making it accessible again, which implies that they are making efforts to address compliance with relevant regulations. 

To our knowledge, the aforementioned data sets represent the only existing collections of human faces specifically intended for training AI models.  Unfortunately, all of them suffer from a variety of drawbacks that diminish their usefulness as benchmarks for FR models: some of them are no longer available due to legal and privacy concerns, and others risk removal for the same reasons. On the other hand, other data sets are entirely synthetic, lack a sufficient number of images to train robust FR models, or are labeled for tasks other than FR. Table \ref{tab: datasets comp} summarizes their characteristics and compares them with the MTF data sets we present in this paper (non-curated and curated).

\begin{table*}[ht]
\small\centering
\caption{Features of the surveyed data sets on human faces. The last two rows show the features of our MTF data sets (non-curated and curated).\textit{}}
\label{tab: datasets comp}
\begin{adjustbox}{width=\textwidth}
\begin{tabular}{@{}lccccc@{}}
\hline
{\em Data set}        & \begin{tabular}[c]{@{}c@{}}{\em Images}\end{tabular} & {\em Annotations}                                                                           & {\em Accessibility}  & {\em Legal compliance}                                                                                                                                                           & {\em Observations}                                                                                                                                                                          \\ \hline
CMU Face Images & 640                                                         & \begin{tabular}[c]{@{}c@{}}Pose, expression,\\ eyes, size\end{tabular}                & Available     & \begin{tabular}[c]{@{}c@{}}Contains images of the research \\ team with their consent\end{tabular}                                                                 & \begin{tabular}[c]{@{}c@{}}Cannot be used\\ to train FR models\end{tabular}                                                                                            \\ \hline
CelebA data     & 202,599                                                     & \begin{tabular}[c]{@{}c@{}}5 landmark locations, \\ 40 binary attributes\end{tabular} & Available     & \begin{tabular}[c]{@{}c@{}}Published under the Chinese \\ copyright law\end{tabular}                                                                               & \begin{tabular}[c]{@{}c@{}}Images collected regardless \\ of their original copyright.\\ Cannot be used\\ to train FR models\end{tabular}                          \\ \hline
DigiFace-1M     & 1,220,000                                                   & Identities                                                                     & Available     & \begin{tabular}[c]{@{}c@{}}Synthetic images, \\ no copyright \\ laws apply\end{tabular}                                                                        & \begin{tabular}[c]{@{}c@{}}Can be used only to train FR models.\\ The synthetic nature limits the \\ data's representativeness\end{tabular}                                           \\ \hline
LFW             & 13,233                                                      & 5,749 identities                                                                      & Available     & \begin{tabular}[c]{@{}c@{}}Created from images \\ with Creative Commons \\ copyright licenses\end{tabular}                                                         & \begin{tabular}[c]{@{}c@{}}The data cannot be used before\\ being processed and cleaned\end{tabular}                                                                                 \\ \hline
Megaface        & 1,000,000                                                   & \begin{tabular}[c]{@{}c@{}}Face location\\ in the image\end{tabular}                  & Available     & \begin{tabular}[c]{@{}c@{}}Created from images \\ with Creative Commons \\ copyright licenses\end{tabular}                                                         & \begin{tabular}[c]{@{}c@{}}Cannot be used\\ to train FR models\end{tabular}                                                                                            \\ \hline
WIDER FACE      & 32,203                                                      & \begin{tabular}[c]{@{}c@{}}Face location\\ in the image\end{tabular}                  & Available     & \begin{tabular}[c]{@{}c@{}}Ethical \\ collection practices\end{tabular}                                                                                   & \begin{tabular}[c]{@{}c@{}}Cannot be used\\ to train FR models\end{tabular}                                                                                            \\ \hline
MALF            & 5,250                                                       & \begin{tabular}[c]{@{}c@{}}Face location\\ in the image\end{tabular}                  & Available     & \begin{tabular}[c]{@{}c@{}}Ethical \\ collection practices\end{tabular}                                                                                   & \begin{tabular}[c]{@{}c@{}}Cannot be used\\ to train FR models\end{tabular}                                                                                            \\\hline
IMDB-WIKI       & 523,051                                                     & Age estimation                                                                        & Available     & \begin{tabular}[c]{@{}c@{}}Collected from\\ IMDB and Wikipedia\end{tabular}                                                                                        & \begin{tabular}[c]{@{}c@{}}Cannot be used\\ to train FR models\end{tabular}                                                                                            \\\hline
FFHQ            & 70,000                                                      & N/A                                                                                   & Available     & \begin{tabular}[c]{@{}c@{}}Downloaded from\\  Flickr, they are \\ subject to the Creative\\  Common copyright licenses\end{tabular}                            & \begin{tabular}[c]{@{}c@{}}Cannot be used\\ to train FR models.\\ Can only aid in the \\ development of generative \\ adversarial networks\end{tabular} \\\hline
VGGFace2        & 3,300,00                                                    & 9,000 identities                                                                      & Not available & \begin{tabular}[c]{@{}c@{}}Downloaded from Google \\ image search without taking\\  into account the copyright \\ ownership of the images\end{tabular}        &  \begin{tabular}[c]{@{}c@{}}Not accessible due to legal\\  and ethical implications\end{tabular}                                                                                                                                                                                      \\\hline
UMDFaces        & 367,888                                                     & \begin{tabular}[c]{@{}c@{}}8,277 identities,\\ 21 biometric keypoints\end{tabular}    & Not available & No information is given                                                                                                                                           &    \begin{tabular}[c]{@{}c@{}}Not accessible due to legal\\  and ethical implications\end{tabular}                                                                                                                                                                                    \\\hline
\emph{Non-curated  MTF}   & 132,816                                                     & \begin{tabular}[c]{@{}c@{}}640 identities, age,\\ gender, race\end{tabular}                & Available     & \begin{tabular}[c]{@{}c@{}}Ensuring ethical collection \\ practices, created from images \\ with Creative Commons or\\ public-domain copyright licenses\end{tabular} & \begin{tabular}[c]{@{}c@{}} Automatically labeled \\and processed\\\end{tabular}                                                           \\\hline
\emph{Curated  MTF}       &  5,246                                                           & \begin{tabular}[c]{@{}c@{}}240 identities, age,\\ gender, race\end{tabular}                & Available     & \begin{tabular}[c]{@{}c@{}}Ensuring ethical collection \\ practices, created from images \\ with Creative Commons or \\public-domain copyright licenses\end{tabular} &      \begin{tabular}[c]{@{}c@{}} Manually labeled \\and processed  \end{tabular}                    \\ \hline                                                                                                                                                          
\end{tabular}
\end{adjustbox}
\end{table*}

\section{Data collection, processing, and labeling}
\label{sec: methodologyu}

The MTF data set has been carefully collected to take advantage of the above-mentioned exception in Article 9 of the GDPR, which allows for the collection and processing of personal data that have been voluntarily made public by the data subject/owner. Along this line, our data collection process prioritized privacy by exclusively focusing on publicly known individuals.
This ensures the legal basis needed for publicly releasing the data set and hence its availability and longevity. Furthermore, we sought and obtained approval from the Board of Operational Ethical and Legality Evaluation (BOEL) of the SoBigData++ H2020 project. 

This section provides an overview of the methodologies employed for selecting, downloading, processing, and labeling the images in the non-curated and curated MTF data sets.

\subsection{Data collection}

The initial phase of our data collection process involved selecting the celebrities to be included in the data set. We utilized the IMDB
website to search for celebrity names across different regions. To enhance the diversity and inclusivity of the data sets, we intentionally included four distinct ethnicities, which are consistent with those employed by the United States Census Bureau:
Asian (Chinese/Korean), Asian (Indian), Black, and White.

In order to maintain a balanced and realistic distribution, we included an equal number of male and female individuals. This approach ensures fairness and avoids any gender bias within the collected data.
Recognizing age as a crucial factor, we also sought to include an equal number of young and old celebrities in the data sets. This consideration adds another dimension to the data and enables research related to age-based analysis.

To distinguish between young and old identities within both MTF data sets, we labeled as young people between ages 18 and 49, and as old people aged 50 or older.

The search we conducted on the IMDB website resulted in the selection of an equal number of celebrities from each of the four ethnicities considered. Our selection process relied on the published lists available on the IMDB website. We conducted searches based on specific criteria and identified celebrities who appeared on these lists. The selection was made according to their presence and ranking on these lists.

The distribution of the selected celebrities from each ethnic group is as follows: 40 old male celebrities, 40 old female celebrities, 40 young male celebrities, and 40 young female celebrities. This totals  160 celebrities selected from each of the four ethnicities, resulting in a grand total of 640 distinct identities. This systematic approach ensures an initial balanced representation within each ethnicity and across different age and gender groups.

To download the images of the chosen celebrities, we used the icrawler \footnote{\url{https://pypi.org/project/icrawler/}} 
library. For this purpose, we chose the Bing
search engine as a source of the images due to its advanced filtering capabilities by image copyright.
Bing offers six different options: ``Creative Commons'', ``public domain'', ``free to share and use non-commercially'', ``free to share and use commercially'', ``free to modify, share, and use non-commercially'', and ``free to modify, share, and use commercially''. In contrast, other search engines such as Google only provide two options: ``Creative Commons'' without any additional specifications, and ``commercial and others''.

The icrawler library proved to be an excellent resource as it provided us with the ability to select a copyright filter for the downloaded images employing the filter options offered by the search engine. By leveraging this functionality, we were able to narrow down our image selection to those aligned with our desired copyright requirements. Specifically, we focus on images that fall under the public domain or Creative Commons licenses, granting us the rights to modify, share, and use the images commercially.

To compile a comprehensive data set, we chose not to impose a limit on the number of images downloaded per identity due to the limited availability of images with the desired copyright licenses. The image crawling process stopped once no further images were obtainable.

At the end of this phase, the data set consisted of a total of $204,712$ images.
However, the distribution of these images among the various identities was not uniform. This uneven distribution was primarily influenced by two factors: the availability of images on the internet for each celebrity and the individual choices made by celebrities regarding copyright permissions. As a result, the data set exhibited non-iidness (nonidentical and independent distribution among identities). Nevertheless, this non-iidness accurately reflected the true distribution of data available online for those specific celebrities under the required licenses.

\subsection{Automatic processing}
\label{processing}

To ensure accurate training of AI models for facial image classification, identity recognition, and human face generative models, it is crucial to have facial images that focus on the faces themselves, while minimizing background details. Therefore, a necessary pre-processing step involves cropping the faces from the crawled images. 

To this end,
each downloaded image was processed through the Haar cascades method \cite{viola2001rapid}, a machine learning-based approach that involves training a cascade function with input data to automatically detect facial regions within the images. For this purpose, we leveraged the pre-trained Haar cascades method available in the OpenCV
\cite{mordvintsev2014opencv} library. This automated process proved efficient in providing the bounding boxes of the facial areas in the images. These boxes varied in size depending on the appearance of each face in the original images. We then utilized these bounding boxes to crop and isolate the facial regions from the acquired images. As a result, we obtained a collection of $132,816$ cropped images, representing $640$ distinct identities. 

Note that the total number of images was lower than that of the initially crawled images mainly due to the presence of mismatches caused by the search engine. Some of the crawled images were not relevant to the desired facial recognition task and did not even contain any recognizable faces.
It is also worth mentioning that the cropped images had varying resolutions according to the size of their respective original images.
To make the images more amenable to AI model training, they underwent a final automated process to standardize their sizes with techniques similar to those used in previous data sets like LFW, WIDER FACE, and MALF. As a result, all images were resized to a uniform resolution of 1024 x 1024 pixels.
\emph{The resulting images constitute the non-curated MTF data set}.

\subsection{Manual processing}
\label{processing-curated}

At this stage, the images were manually processed in order to build a curated version of the MTF data set focusing on maximizing classification quality. 
Three experienced human evaluators were involved to visually analyze each cropped image. Their primary objective was to confirm the presence of the desired celebrity's face in each image. This thorough examination process encompassed the following steps:

\begin{itemize}

    \item First, the human experts filtered the data set by eliminating each face image that did not belong to the correct identity. 
    For instance, many of the crawled images featured groups of celebrities or celebrities attending festivals with a large number of fans in the background, leading to multiple faces being cropped from the same image. However, only one of those faces belonged to the desired celebrity, which required the removal of the rest. Additionally, search engine results occasionally included images of celebrity relatives, such as spouses, children, or co-workers, which were also excluded by the experts during this stage.

    \item Second, to ensure the suitability of the data set for training robust AI models, the experts implemented several further filtering criteria. Images that contained hidden or too dark parts of the celebrity's face, such as sunglasses or hands covering the mouth or eyes, were removed. In addition, images that were hand drawn, artificially altered, or generated by AI algorithms were excluded. The data set was further refined by eliminating images where the celebrity's face appeared unnatural due to artistic effects or intrusive movie-related makeup, typically seen in images taken during movie shoots. Furthermore, images with visual disturbances such as heavy pixelation or blurring, which could potentially mislead the AI model, were also excluded from the data set. These filtering criteria ensured that the data set consists only of high-quality, natural, and unaltered facial images suitable for training AI models.

    \item Third, to mitigate the risk of data leakage and avoid any potential additional unnecessary computational cost during AI model training, the experts implemented measures to identify and remove duplicate or highly similar images (e.g., burst shoots) from the data set.
    They employed three techniques for this purpose. The first technique involved an automated code written using the OpenCV library. This code calculated the similarity between images and alerted the experts about potential duplicates. The second technique utilized the Duplicate Photos Fixer
    \footnote{\url{https://www.duplicatephotosfixer.com}} 
    application, which systematically scanned the data set and compiled a list of potentially duplicated or very similar images. Lastly, the experts visually inspected the data set to ensure that no very similar or duplicate images were overlooked. This manual examination served as a final verification step to confirm the absence of any remaining duplicates.  
    
    \item Fourth, to ensure the usefulness of the data set in different tasks, experts eliminated images that did not meet the criteria for all desired tasks. For instance, images depicting outdated representations of currently old individuals that could impact the age classification task, or images of the celebrities during their childhood that could affect both face recognition and age classification, were removed. 
  
\end{itemize}

Finally, the experts conducted a thorough examination of the remaining images to assess their usefulness. During this stage, the experts counted the number of images for each identity that remained in the data set after undergoing all the aforementioned processes. Individuals with less than five remaining images were excluded. This step was essential because a limited number of images would not provide sufficient data for effective training, validation, and testing on the face recognition task. After this meticulous manual curation, the size of the data set was reduced to $5,246$ images and $240$ identities, but each had a sufficient amount of data available. \emph{These images constitute the curated MTF data set}.

Although this manual curation resulted in a significant reduction in the number of images, it is important to note that this filtering process is a clear advantage over other data sets such as Megaface, VGGFace2, and UMDFaces, which include a large number of systematically crawled internet images that can introduce significant noise during training.

\subsection{Data labeling}

After the data processing phase, the images were labeled based on the predetermined criteria established during the collection phase.
These labels were specifically tailored for the four tasks outlined below: 
\begin{enumerate}
    \item \emph{Face recognition}: For the non-curated MTF data set, each image was assigned one of 640 identities they were crawled from. In the case of the curated MTF data set, each image was assigned one of the 240 remaining identities.
    \item \emph{Race classification}: Each image was assigned one of the following four labels according to the race of the corresponding individual:
    \begin{itemize}
        \item Asian (Chinese/Korean) 
        \item Asian (Indian) 
        \item Black
        \item White
    \end{itemize}
    \item \emph{Gender classification}: Each image was assigned one of the two following labels: \begin{itemize}
        \item Male
        \item Female
    \end{itemize}
    \item \emph{Age classification}: Due to images corresponding to different ages of the corresponding individual, we treated this task as a binary classification. Each of the images was assigned one of the following two labels: \begin{itemize}
        \item Young
        \item Old
    \end{itemize}

\end{enumerate}

\begin{figure}[ht]
\centering
\includegraphics[width=0.48\textwidth]{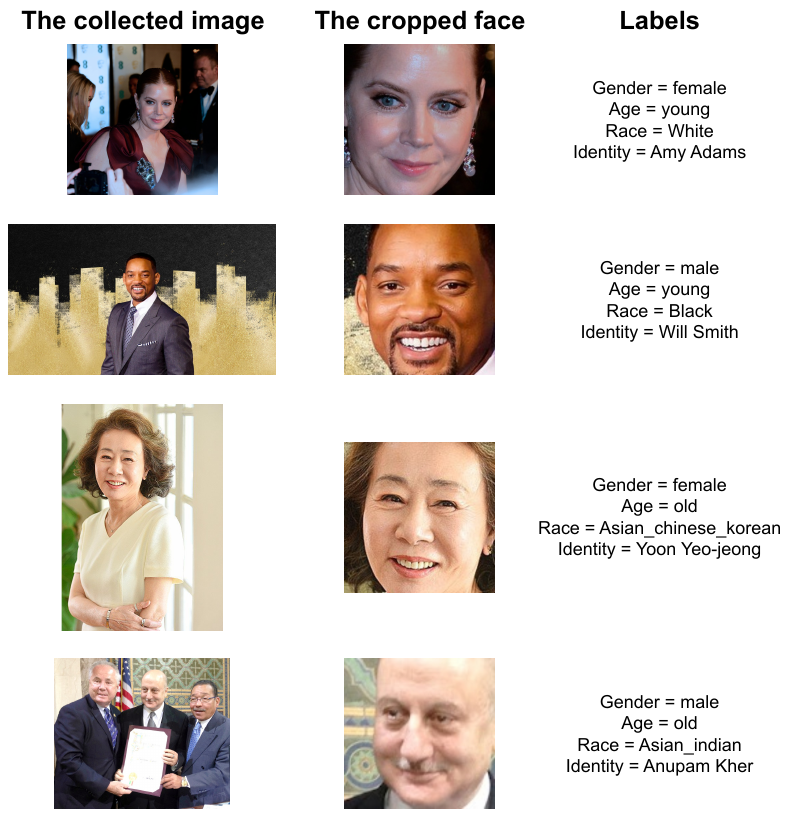}
\caption{Examples of collected, processed, and labeled images from the MTF data sets}
\label{fig: collection_result}
\end{figure}

Fig. \ref{fig: collection_result} illustrates the procedures explained in this section, by showing the transition from the original collected images to their corresponding cropped versions and the assigned labels.

\section{The Multi-Task Faces (MTF) data sets}
\label{sec: data Specification}

After completion of the processing and labeling of all images, the distribution of identities assigned to each label in the four different tasks was shown in Table \ref{tab: identities labels}. While our initial efforts aimed to crawl a balanced number of images across all tasks and labels, the actual distribution of the data available online resulted in an imbalance within the curated data set. This imbalance can be attributed to several factors: i) celebrities from different regions of the world publish their images at different rates and under different copyright licenses; ii) young celebrities tend to publish their images more frequently than old celebrities; and iii) old celebrities often have more images from their younger days than from their current age.

\begin{table}[ht]
\small\centering
\caption{Distribution of the identities and images among the different tasks of both data sets}
\label{tab: identities labels}
\begin{adjustbox}{width=0.48\textwidth}

\begin{tabular}{@{}lc|cc|cc@{}}
\hline
\multirow{2}{*}{Task}                  & \multirow{2}{*}{Label}                & \multicolumn{2}{c}{Curated MTF} & \multicolumn{2}{|c}{Non-curated MTF}    \\ 
                                                                               && Identities       & Images & Identities       & Images \\\hline
\begin{tabular}[c]{@{}l@{}}Face \\ recognition\end{tabular}                       & Identities                                                         & 240               & 5,246       & 640                             & 132,816   \\ \hline
\multirow{4}{*}{\begin{tabular}[c]{@{}l@{}}Race \\ classification\end{tabular}}   & \begin{tabular}[c]{@{}l@{}}Asian\\ (Chinese\\ /Korean)\end{tabular} & 80                & 1,715       & 160                             & 32,257   \\
                                                                                  & \begin{tabular}[c]{@{}l@{}}Asian\\ (Indian)\end{tabular}            & 49                & 820         & 160                             & 31,503    \\
                                                                                  & Black                                                               & 35                & 478         & 160                             & 31,612    \\
                                                                                  & White                                                               & 76                & 2,133       & 160                             & 37,444   \\ \hline
\multirow{2}{*}{\begin{tabular}[c]{@{}l@{}}Gender \\ classification\end{tabular}} & Males                                                               & 130               & 2,490       & 320                             & 68,702  \\
                                                                                  & Females                                                             & 110               & 2,756       & 320                             & 64,114   \\ \hline
\multirow{2}{*}{\begin{tabular}[c]{@{}l@{}}Age \\ classification\end{tabular}}    & Young                                                              & 190               & 4,682       & 320                             & 67,462   \\
                                                                                  & Old                                                                 & 50                & 514         & 320                             & 65,354   \\ \hline
\end{tabular}
\end{adjustbox}
\end{table}

The four classification tasks are remarkably diverse.
\begin{itemize}
\item The FR task involves a classification problem with many categories, as it requires labeling identities based on their names. 
\item The second is the RC task, which presents a multi-label classification problem with four labels. The distribution of identities in this task is imbalanced for the curated data set, as it features two majority groups and two minority groups. The majority groups consist of Asians (Chinese/Korean) and Whites, as celebrities from these regions tend to share their images with our target copyright licenses more frequently than celebrities from the other two regions, namely Asian (Indian) and Black celebrities.
\item The GC task represents the first binary classification problem. The data are relatively well balanced between the two labels in the curated data set, with slightly more male identities present. However, both male and female identities share their images at a similar rate, although females tend to share a higher number of images overall.
\item Finally, the AC task also represents a binary classification problem. In contrast to the GC task, the AC task exhibits an extremely unbalanced data distribution in the curated data set. This disparity is due to the reasons listed above, resulting in the majority of identities and images falling under the ``young'' category, while the ``old'' category comprises only 514 images that belong to a mere 50 identities. 

\end{itemize}

These demographic imbalances, particularly those that involve underrepresented age and racial groups, have important implications for fairness. Models trained on skewed datasets may exhibit biased performance, favoring majority groups while underperforming on minority ones. This can lead to unfair or unreliable outcomes when such models are deployed in real-world applications. Consequently, researchers using the MTF data sets are encouraged to carefully consider these imbalances, especially when evaluating or reporting model performance, and to apply bias mitigation techniques or rebalancing the data where necessary. Note that {\em if users of the curated MTF data set wish to rebalance
the distribution within each task, they can do so by selecting balanced subsets of the data set.}

Moreover, the MTF data sets offer potential for a wide range of tasks that extend beyond the four previously mentioned. Their flexibility allows defining multi-criteria classification tasks involving two or even three attributes, which provides diverse options for AI model development. In Table \ref{tab: identities intersection}, we present the count of identities and the respective count of images for the cross-tabulation of race and gender, as well as the cross-tabulation of race and age for the curated MTF data set.

\begin{table}[ht]
\centering

\caption{Counts of identities and images when cross-tabulating by race and gender, and by race and age in the curated MTF data set. The first figure in each cell refers to the count of identities, and the second figure to the count of images.}
\label{tab: identities intersection}%
\begin{tabular}{@{}ll|cc|cc@{}}
\hline
\multicolumn{2}{@{}c@{}}{\multirow{2}{*}{Labels}}                                                                                                                   & \multicolumn{2}{|c}{Gender labels}     & \multicolumn{2}{|c}{Age labels}       \\ 
\multicolumn{2}{@{}c@{}}{}                                                                                                                                          & \multicolumn{1}{|c}{Males}   & Females & \multicolumn{1}{@{}c@{}}{Young}   & Old    \\ \hline
\multicolumn{1}{@{}c@{}}{\multirow{4}{*}{\begin{tabular}[c]{@{}l@{}}Race\\ labels\end{tabular}}} & \begin{tabular}[c]{@{}l@{}}Asian\\ (Chinese\\/Korean)\end{tabular} & \multicolumn{1}{@{}c@{}}{36/1,020} & 44/695  & \multicolumn{1}{@{}c@{}}{69/1,612} & 11/103 \\ 
\multicolumn{1}{@{}c@{}}{}                                                                       & \begin{tabular}[c]{@{}l@{}}Asian \\ (Indian)\end{tabular}        & \multicolumn{1}{@{}c@{}}{22/300}  & 27/520  & \multicolumn{1}{@{}c@{}}{43/749}  & 6/71   \\ 
\multicolumn{1}{@{}c@{}}{}                                                                       & Black                                                            & \multicolumn{1}{@{}c@{}}{18/194}  & 17/384  & \multicolumn{1}{@{}c@{}}{30/533}  & 5/45   \\ 
\multicolumn{1}{@{}c@{}}{}                                                                       & White                                                            & \multicolumn{1}{@{}c@{}}{54/976}  & 22/1,157 & \multicolumn{1}{@{}c@{}}{48/1,788} & 28/345 \\ \hline
\end{tabular}
\end{table}

Similarly, Table \ref{tab: identities intersection 2} reports the number of identities and their corresponding images for the cross-tabulation of age and gender in the curated data set.  

\begin{table}[ht]
\centering
\caption{Counts of identities and images when cross-tabulating by age and gender in the curated MTF data set. The first figure in each cell refers to the count of identities, and the second figure to the count of images.}
\label{tab: identities intersection 2}%
\begin{tabular}{@{}llcc@{}}
\hline
\multicolumn{2}{@{}c@{}}{\multirow{2}{*}{Labels}}                                                        & \multicolumn{2}{@{}c@{}}{Gender label}      \\ 
\multicolumn{2}{@{}c@{}}{}                                                                               & \multicolumn{1}{@{}c@{}}{Males}   & Females \\ \hline
\multicolumn{1}{@{}c@{}}{\multirow{2}{*}{\begin{tabular}[c]{@{}c@{}}Age \\ labels\end{tabular}}} & Young & \multicolumn{1}{@{}c@{}}{92/2,049} & 98/2,633 \\ 
\multicolumn{1}{@{}c@{}}{}                                                                       & Old   & \multicolumn{1}{@{}c@{}}{38/441}  & 12/123  \\ \hline
\end{tabular}
\end{table}

\subsection{Curated data splitting for training and testing}

As the main purpose of the curated MTF data set is to train and test classification models, we performed two rounds of data splitting.
In the first split, we allocated $70\%$ of the curated data ($3,662$ images) for training purposes, while the remaining $30\%$ was set aside for validation and testing. Then we took a second split on the $30\%$ portion, which we divided into two subsets: approximately $20\%$ of this portion ($332$ images) was designated as validation data, to be employed for fine-tuning and optimization of the model. The remaining $80\%$ ($1,252$ images) were allocated as test data, to be used as an independent evaluation set to assess the generalization and accuracy of the trained model.
To facilitate a consistent evaluation in future experiments conducted on the data set, we maintained the same data splitting across all tasks. 
We did so following a methodology that ensured that each subset of the data contained at least one image that represents each label within each task. Specifically, 
we made sure to include at least one image per label in each of the validation and test subsets, as well as a minimum of three images per label in the training subset. This approach was particularly important for the FR task, where the data is more spread across the 240 identity labels, which requires a more evenly distributed data set.

\subsection{Data release}

Both MTF data sets have been made available under the Creative Commons copyright license, granting permissions for sharing, modifying, and commercial usage. This respects the copyright of the original images. Users are obligated to provide proper attribution to the authors of the data sets and the owners of the original images when utilizing any of the MTF data 
sets. This includes citing the present paper and specifying the source of the data.
 
\begin{figure}[ht]
\centering
\includegraphics[width=0.48\textwidth]{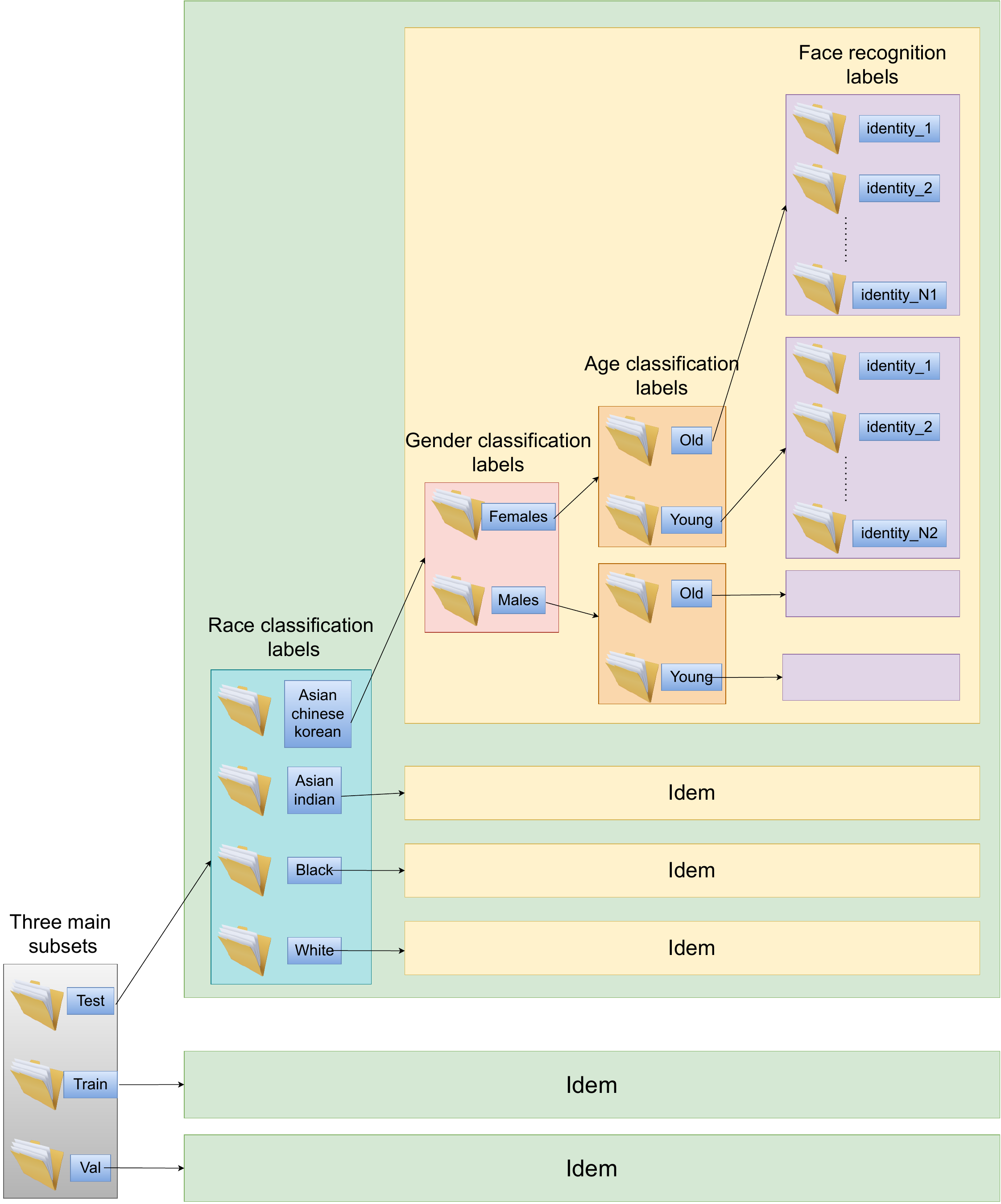}
\caption{Organization of the folders in the released data set}
\label{fig: data Folder}
\end{figure}

The curated MTF data set has been made accessible through the European SoBigData++ catalogue. It can also be accessed alongside the non-curated data set using the following GitHub webpage that points to the catalogue: \url{https://github.com/RamiHaf/MTF_data_set}

The release includes the full labeling of images by organizing them in an appropriate folder structure depicted in Fig. \ref{fig: data Folder}. 
The main folder of the curated data set comprises three subsets,`Train', `Val', and `Test' to be used, respectively, for AI model training, 
AI model validation and hyperparameter tuning, 
and AI model testing (evaluation of the trained model).
Inside each subset, which is also the root folder of the non-curated data set, there are four folders corresponding to race classification denoted as `Asian\_chinese\_korean', `Asian\_indian', `Black', and `White'. Within each race folder, there are two additional folders named `Males' and `Females', representing the two labels used for gender classification.
Inside each of the gender folders, there are two more folders labeled `Young' and `Old', which indicate the two age categories used for age classification.
Within each `Age' folder, there are folders corresponding to the identities. 

With this structure, researchers have the flexibility to rearrange the data within the data sets according to their specific preferences and requirements, e.g., by defining multi-criteria classification tasks involving several attributes as mentioned in Tables \ref{tab: identities intersection} and \ref{tab: identities intersection 2}.

Additional items made available through the above-mentioned GitHub webpage are as follows:

\begin{itemize}
\item To facilitate and support future research on the MTF data, we have released a second curated version that is pre-split 
and organized specifically for users
interested in only one of the four tasks. This version includes four separate folders: `race\_classification', `gender\_classification', `face\_recognition', and `age\_classification'. Each folder contains all the images that are labeled according to the corresponding task.  
\item We release also the trained models we evaluate in the next section. Since these models provide baseline results for the different tasks supported by the curated data set, by releasing them we aim at facilitating future investigations conducted on this data. 
\item Finally, we release the Python code we used for evaluating 
both data sets (non-curated and curated) with the various DL models mentioned in the next section. This code can serve as a baseline for other researchers to 
compare their work with the results reported here.
\end{itemize}

\section{Experimental results}
\label{sec: results}

To assess the efficacy of the curated MTF data set for training DL models and establish a performance baseline across its supported tasks, we conducted evaluations using five well-known DL classification models. In the following we describe the evaluation metrics employed to assess the performance achieved, the specific models used in the experiments, and the obtained results.

\subsection{Evaluation metrics}
We employed the following standard evaluation metrics to measure the performance of DL models trained on the MTF data set:
\begin{itemize}
\item  \emph{Accuracy}: Ratio of correctly predicted instances to the total number of predicted instances in the data set. It provides an overall measure of the classification correctness:

\begin{equation*}
    Accuracy = \frac{Number\_of\_correct\_predictions}{Total\_number\_of\_predictions}.
\end{equation*}

\item \emph{Precision}: Proportion of true positive predictions to the sum of true positive and false positive predictions. It indicates the accuracy of positive predictions:

\begin{equation*}
    Precision = \frac{True\_positives}{True\_positive + false\_positive}.
\end{equation*}

\item \emph{Recall}: Proportion of true positive predictions to the sum of true positive and false negative predictions. It captures the model's ability to identify positive instances:

\begin{equation*}
    Recall = \frac{True\_positives}{True\_positive + false\_negative}.
\end{equation*}

\item \emph{F1 Score}: Harmonic mean of precision and recall. It offers a balanced measure of the model's accuracy:
\begin{equation*}
    F1 = 2 \frac{1}{\frac{1}{Precision}+\frac{1}{Recall}}.
\end{equation*}

\end{itemize}

\subsection{DL models and hyperparameters}

To establish a baseline for evaluating the performance of the curated MTF data set, we performed assessments using five well-known DL classification models. These models vary in depth and complexity. Table \ref{Tab: models} provides an overview of the models chosen, including their number of parameters and the year they were published. Furthermore, the table summarizes the main purpose behind their design and highlights their advantages at time of their release.

\begin{table}[ht]

\caption{DL models used for evaluation}
\label{Tab: models}
\centering

\begin{adjustbox}{width=0.48\textwidth}
\begin{tabular}{@{}llll@{}}
\hline
Models                                 & Parameters  & Year & Purpose                                                                                                                                                                                     \\ \hline\hline
MobileNetV3 \cite{howard2019searching} & 4,509,472   & 2019 & \begin{tabular}[c]{@{}l@{}}To run in embedded systems.\\ Reduced number of parameters.\end{tabular}                                                                                           \\ \hline
AlexNet \cite{krizhevsky2012imagenet}  & 57,987,120  & 2012 & \begin{tabular}[c]{@{}l@{}}Extremely capable DL model at\\ that time, to be trained on the \\ImageNet \cite{deng2009imagenet} data set.\end{tabular}                                           \\ \hline
ResNet50 \cite{he2016deep}             & 23,999,792  & 2016 & \begin{tabular}[c]{@{}l@{}}Reduced number of parameters \\and matrix multiplications.\\ Enables much faster training of\\ each layer. Uses a stack of three\\ layers rather than two.\end{tabular} \\ \hline
VGG16 \cite{simonyan2014very}          & 135,243,824 & 2014 & \begin{tabular}[c]{@{}l@{}}Presents the small (3 × 3)\\ convolution filters. Introduces the \\deep network with 16 layers.\end{tabular}                                                         \\ \hline
ConvNeXT \cite{liu2022convnet}         & 87,812,464  & 2022 & \begin{tabular}[c]{@{}l@{}}Introduces the Vision Transformer\\ into image processing  through \\depthwise convolution.\end{tabular}     \\        \hline                          
\end{tabular}
\end{adjustbox}
\end{table}

To ensure a fair evaluation across the different models, we used the same hyperparameters during both training and evaluation. However, a minor exception was made for the batch size. Due to resource limitations, we adjusted the batch size to be smaller for the models with a large number of parameters, allowing them to run efficiently on the available hardware. Conversely, for the smaller models, we increased the batch size to optimize time efficiency. This approach enabled us to conduct a comprehensive and equitable assessment of the performance of the models.

Specifically, the chosen hyperparameters included the cross-entropy loss function \cite{zhang2018generalized}, the Adam optimizer \cite{kingma2014adam}, and an automated learning rate scheduler \cite{kim2021automated}. The initial learning rate was set at 0.001, and it decreased by a factor of 0.1 every 20 epochs. The training process was limited to a maximum of 100 epochs, with early stopping implemented if the training loss did not improve for five consecutive epochs. 

The batch size for the three smaller models was set to 128 images, while for the two larger models it was reduced to 32 images.

The experiments were carried out on a computer equipped with an AMD Ryzen 5 3600 CPU running at a base speed of 3.6 GHz, 32 GB of RAM, and an NVIDIA GeForce RTX 3060 GPU with 12 GB of dedicated RAM.

\subsection{Results}

We first report on the performance of the five DL models on the curated MTF data set. All experiments were conducted using the predefined models available in the PyTorch.
\cite{NEURIPS2019_9015} library. For each of the four tasks of the data set, we report the results obtained from random guessing, the performance of the five models when trained from scratch on the data set, and the performance of the five models when fine-tuned on the curated MTF data set using pre-trained weights provided by the PyTorch library. These pre-trained weights were obtained by training the model using the ImageNet data set \cite{deng2009imagenet}.

\subsubsection{Face recognition task}

Table \ref{Tab: FR_results} reports the performance of DL models on the curated MTF data set for the face recognition task, which comprises 240 labels.

\begin{table}[ht]
\centering
\caption{Performance of the DL models on the curated MTF data set for the face recognition task. Boldface figures are the best in each column. S denotes that the model is trained from scratch, while P denotes that the model was pre-trained.}
\begin{adjustbox}{width=0.48\textwidth}
\label{Tab: FR_results}
\begin{tabular}{@{}llrrrrc@{}}
\hline
Model                          & Origin     & Accuracy         & Precision        & Recall           & F1 score          & \begin{tabular}[c]{@{}c@{}}Training time\\ (minutes)/\ \\ number \\ of epochs\end{tabular} \\ \hline\hline
Random guess                   & N/A          & 0.16\%           & 0.00\%           & 0.42\%           & 0.00\%           & N/A                                                                                                \\ \hline
\multirow{2}{*}{MobileNet\_v3} & S & 2.88\%           & 0.01\%           & 0.42\%           & 0.02\%                                                                   & \textbf{94.77} / 70                                                    \\  
                               & P  & 57.35\%          & 49.77\%          & 46.16\%          & 43.88\%                                                                 & 94.99 /71                                                            \\ \hline
\multirow{2}{*}{AlexNet}       & S & 18.93\%          & 8.71\%           & 10.76\%          & 7.97\%                                                                   & 95.56  /76                                                           \\  
                               & P  & 36.66\%          & 27.34\%          & 25.81\%          & 23.94\%                                                               & 101.96  / 82                                                          \\ \hline
\multirow{2}{*}{ResNet 50}     & S & 9.58\%           & 3.13\%           & 4.30\%           & 3.13\%                                                                   & 168.24 / 100                                                           \\  
                               & P  & 58.39\%          & 46.94\%          & 46.01\%          & 44.61\%                                                                & 151.95  / 94                                                          \\ \hline
\multirow{2}{*}{VGG16}         & S & 23.96\%          & 17.89\%          & 16.08\%          & 14.57\%                                                                    & 184.13 / 70                                                           \\  
                               & P  & 34.66\%          & 24.21\%          & 23.63\%          & 21.23\%                                                                & 156.77 / 74                                                           \\ \hline
\multirow{2}{*}{ConvNeXT}      & S & 13.26\%          & 5.13\%           & 6.60\%           & 4.60\%                                                                  & 216.9  / 71                                                           \\  
                               & P  & \textbf{79.87\%} & \textbf{76.29\%} & \textbf{73.31\%} & \textbf{74.07\%}                                                        & 201.74   / 76                                                         \\ \hline
\end{tabular}
\end{adjustbox}
\end{table}

  As expected, all trained models achieved better results than random guess. However,  the MobileNet\_v3 model trained from scratch exhibited subpar performance, just slightly above the random baseline   (accuracy 2.88\% and F1 score 0.02\%). Remarkably, its low recall of just 0.42\% indicates that this model predicted all images on a single label, suggesting that it was not able to accurately extract image features for precise predictions. This outcome for MobileNet\_v3 was not surprising because this model was designed for simple and fast training in embedded systems.
 The other models trained from scratch exhibited better performance than the aforementioned model, with higher recall scores implying predictions across multiple labels and potentially benefiting from increased training epochs and higher learning rates.

 As anticipated, all fine-tuned models with pre-trained weights outperformed their scratch-trained counterparts. This can be attributed to the pre-trained layers of the models, which are adept at identifying important image features and passing them to the model's classifier. 
The recall scores of the pre-trained models indicate successful predictions across multiple correct labels. Taking into account the uneven distribution of the data, the ConvNeXT model stood out with high results, with an accuracy of 79.78\% and an F1 score of 74.07\% due to its novelty and ability to tackle complex tasks using the Vision Transformer. 
 Despite this model showing high performance on the FR task, the complexity of this highly multidimensional classification challenge presents significant difficulties for all other models, including the ConvNeXT model when trained from scratch. As a result, the MTF data set emerges as a valuable asset for encouraging and supporting future research in this area.

 When examining the runtime required to train the models, we observe that each pre-trained model exhibits a very similar training duration if compared to its counterpart trained from scratch. The slight discrepancy in training times can be attributed to the variation in the number of epochs required to achieve the final model. The training time is contingent on the complexity of the model and the number of parameters.
 The longer training times of ResNet50 and ConvNeXT are due to the former using residual layers and the latter employing Vision Transformers, which contribute to their higher complexity. 
 Furthermore, the smaller batch size used in the training of VGG16 and ConvNeXT played a role in the extended training time required by these models.

\subsubsection{Race classification}

The race classification task involves a four-label classification objective. The relative differences across models follow what was observed in the previous FR task.

\begin{table}[ht]
\centering
\caption{Performance of the DL models on the curated MTF data set for the race classification task. Boldface figures are the best in each column. S denotes that the model is trained from scratch, while P denotes that the model was pre-trained.}
\begin{adjustbox}{width=0.48\textwidth}
\label{Tab: RC_results}
\begin{tabular}{@{}llrrrrc@{}}
\hline

Model                          & Origin     & Accuracy         & Precision        & Recall           & F1 score         & \begin{tabular}[c]{@{}c@{}}Training time\\ (minutes)/\ \\ number \\ of epochs\end{tabular}  \\ \hline\hline
Random guess                   & N/A          & 15.34\%          & 3.83\%           & 25.00\%          & 6.65\%           & N/A                                                                                            \\ \hline
\multirow{2}{*}{MobileNet\_v3} & S & 40.97\%          & 10.24\%          & 25.00\%          & 14.53\%                                                                   & \textbf{92.4}  / 70                                                   \\ 
                               & P  & 85.62\%          & 83.12\%          & 81.09\%          & 82.01\%                                                                  & 98.97 / 72                                                            \\ \hline
\multirow{2}{*}{AlexNet}       & S & 67.01\%          & 60.13\%          & 55.31\%          & 56.23\%                                                                 & 92.97     / 73                                                        \\ 
                               & P  & 87.22\%          & 85.82\%          & 84.49\%          & 85.08\%                                                            & 94.22   / 78                                                          \\ \hline
\multirow{2}{*}{ResNet 50}     & S & 60.06\%          & 44.24\%          & 45.46\%          & 43.39\%                                                               & 106.84      / 74                                                      \\ 
                               & P  & 90.18\%          & 88.95\%          & 87.16\%          & 87.98\%                                        & 103.1  / 71                                                           \\ \hline
\multirow{2}{*}{VGG16}         & S & 78.04\%          & 74.65\%          & 71.36\%          & 72.70\%                                                                & 134.27        / 70                                                    \\ 
                               & P  & 92.33\%          & 90.66\%          & 90.95\%          & 90.78\%                                                     & 177.13  / 74                                                          \\ \hline
\multirow{2}{*}{ConvNeXT}      & S & 61.50\%          & 51.40\%          & 48.43\%          & 48.16\%                                                               & 233.86     / 82                                                       \\ 
                               & P  & \textbf{95.77\%} & \textbf{94.41\%} & \textbf{94.94\%} & \textbf{94.65\%}                          & 207.07  / 70                                                          \\ \hline
\end{tabular}
\end{adjustbox}
\end{table}

As shown in Table \ref{Tab: RC_results}, all models demonstrated better performance than the random guess by the end of their training, whether pre-trained or trained from scratch. 
The training times to reach their final form were similar to those shown in Table \ref{Tab: FR_results}. 
The slight differences in training times among tasks were due to varying numbers of training epochs, after which all models reached the stopping condition of non-improved loss for five successive epochs.

The curated data set exhibits an imbalanced nature for this task, with two majority labels: Asian (Chinese/Korean) accounting for 32.38\% of the data, and White comprising 40.97\% of the data. On the other hand, two minority labels are present: Asian (Indian) making up 15.63\% of the data, and Black representing 11.02\% of the data. This imbalance severely affected the MobileNet\_v3 model when trained from scratch: its recall was just 25\%, which indicates that all images were predicted to belong to a single label. That behavior could be attributed to the simplicity of the model and the low number of parameters, which prevented the model from handling such complex data. Moreover, the other models trained from scratch (ResNet\_50, AlexNet, and ConvNeXT) were also affected by the data distribution, with accuracies lower than 70\% and recall scores hovering around 50\%, which suggest they were primarily classifying images into two labels. The exception was the VGG16 model, which achieved a relatively high accuracy of 78.04\% and a recall score of 71.36\%, therefore indicating a successful multi-label image classification.

If we examine the pre-trained models, all of them outperformed their trained-from-scratch counterparts, boasting recall scores over 81\%. This demonstrates their ability to correctly classify images into the four labels and their resilience to the data set imbalance.
The most complex models, ConvNeXT and VGG16, achieved impressively high accuracy and F1 scores, both exceeding 90\%. Their accuracies were  95.77\% and 92.33\%, respectively, with F1 scores of 94.65\% and 90.78\%. 

\subsubsection{Gender classification}

The gender classification task in the curated MTF data set involves a binary classification problem with a well-balanced data distribution: approximately 47.12\% of the images are labeled as males, and 52.88\% are labeled as females in both the training and the test sets. Thus, GC should be an `easy' problem for most DL models, as the results reported in Table \ref{Tab: GC_results} suggest.

\begin{table}[ht]
\centering
\caption{Performance of the DL models on the curated MTF data set for the gender classification task. Boldface figures are the best in each column. S denotes that the model is trained from scratch, while P denotes that the model was pre-trained.}
\begin{adjustbox}{width=0.48\textwidth}
\label{Tab: GC_results}
\begin{tabular}{@{}llrrrrc@{}}
\hline
Model                          & Origin     & Accuracy & Precision & Recall  & F1 score &\begin{tabular}[c]{@{}c@{}}Training time\\ (minutes)/\ \\ number \\ of epochs\end{tabular} \\ \hline\hline
Random guess                   & N/A          & 45.61\%  & 40.60\%   & 43.88\% & 39.18\%  & N/A                                                           \\ \hline
\multirow{2}{*}{MobileNet\_v3} &S & 52.88\%  & 26.44\%   & 50.00\% & 34.59\%                                                          & \textbf{80.3}    / 59                                                       \\ 
                               & P  & 76.60\%  & 77.94\%   & 75.89\% & 75.93\%                                                & 100.74  / 74                                                          \\ \hline
\multirow{2}{*}{AlexNet}       & S & 70.13\%  & 71.35\%   & 69.28\% & 69.05\%                 & 90.33              / 71                                               \\ 
                               & P  & 88.18\%  & 88.16\%   & 88.11\% & 88.13\%                                 & 92.99  / 73                                                           \\ \hline
\multirow{2}{*}{ResNet 50}     & S & 70.85\%  & 70.80\%   & 70.57\% & 70.62\%                                                          & 142.16     / 78                                                       \\ 
                               & P  & 97.60\%  & 97.59\%   & 97.61\% & 97.60\%                                              & 127.24   / 71                                                         \\ \hline
\multirow{2}{*}{VGG16}         & S & 90.65\%  & 90.61\%   & 90.67\% & 90.63\%             & 181.9    / 76                                                         \\ 
                               & P  & 97.28\%  & 97.29\%   & 97.26\% & 97.27\%                                                & 211.4 / 80                                                            \\ \hline
\multirow{2}{*}{ConvNeXT}      & S & 73.80\%  & 73.92\%   & 73.97\% & 73.80\%               & 209.88                 / 71                                           \\ 
                               & P  & \textbf{98.88\%}  & \textbf{98.86\%}   & \textbf{98.90\%} & \textbf{98.88\% }                              & 219.22    / 74                                                        \\ \hline
\end{tabular}
\end{adjustbox}
\end{table}

When MobileNet\_v3 was trained from scratch, it exhibited low performance, as it did in the previous tasks: it achieved an accuracy of 52.88\%, and recall of 50\%, indicating that it labeled all images as females. Thus, the proposed curated MTF data set contains images more complicated than the MobileNet\_v3 model can handle even for the easiest task.

However, the remaining trained-from-scratch models performed better, with accuracies exceeding 70\% and recall scores over 69\%. Among them, VGG16 stood out, achieving impressive results with an accuracy of 90.65\% and a recall of 90.67\%. The complexity of the VGG16 network and its extensive parameter count allowed it to accurately classify the images.

On the other hand, models fine-tuned with pre-trained weights demonstrated superior performance compared to the previously discussed models. Three of these models achieved accuracy above 97\%. Ranging from best to worst accuracy, these models are ConvNeXT,  ResNet 50, and VGG16, with accuracies of 98.88\%, 97.60\%, and 97.28\% respectively, and F1 scores of 98.88\%, 97.60\%, and 97.27\%, respectively. These outstanding results can be attributed to the capabilities of these models, including their large number of parameters and their use of novel layers such as residual and transformer layers.

Regarding the training runtimes, they are comparable to those of the previous tasks for each respective model, taking into consideration the number of training epochs needed for each model to converge, as well as the differences in batch sizes used.

\subsubsection{Age classification}

The binary age classification task in the curated MTF data set exhibits a significant bias, with 89.94\% of the data labeled as `young' and only 10.06\% labeled as `old' in both the training and test sets.

\begin{table}[ht]
\centering
\caption{Performance of the DL models on the curated MTF data set for the age classification task. Bold figures are the best in each column. S denotes that the model is trained from scratch, while P denotes that the model was pre-trained.}
\begin{adjustbox}{width=0.48\textwidth}
\label{Tab: AC_results}
\begin{tabular}{@{}llrrrrc@{}}
\hline
Model                          & Origin     & Accuracy         & Precision        & Recall           & F1 score         & \begin{tabular}[c]{@{}c@{}}Training time\\ (minutes)/\ \\ number \\ of epochs\end{tabular} \\ \hline\hline
Random guess                   & N/A          & 39.06\%          & 51.80\%          & 54.49\%          & 35.27\%          & N/A                                                                                              \\ \hline
\multirow{2}{*}{MobileNet\_v3} & S & 90.01\%          & 44.97\%          & 50.00\%          & 47.35\%                                                                & 101.19  / 72                                                          \\ 
                               & P  & 94.81\%          & 91.06\%          & 78.08\%          & 83.03\%                                                                 & 96.67 / 70                                                            \\ \hline
\multirow{2}{*}{AlexNet}       & S & 90.50\%          & 86.20\%          & 53.48\%          & 54.06\%                                                                & \textbf{94.67}  / 65                                                  \\ 
                               & P & 95.93\%          & 90.71\%          & 85.75\%          & 88.03\%                                                                & 96.86  / 67                                                           \\ \hline
\multirow{2}{*}{ResNet 50}     & S & 90.02\%          & 78.37\%          & 50.75\%          & 48.92\%                                                              & 110.96     / 70                                                       \\ 
                               & P  & 95.85\%          & 89.14\%          & 87.47\%          & 88.28\%                                                                  & 140 / 77                                                              \\ \hline
\multirow{2}{*}{VGG16}         & S & 92.81\%          & 90.26\%          & 66.40\%          & 72.21\%                                                               & 168.09    / 78                                                        \\ 
                               & P  & 96.33\%          & 91.82\%          & 87.03\%          & 89.24\%                                                & 149.55  / 72                                                          \\ \hline
\multirow{2}{*}{ConvNeXT}      & S & 90.50\%          & 86.20\%          & 53.48\%          & 54.06\%                                                            & 211.75    / 75                                                        \\ 
                               & P  & \textbf{97.60\%} & \textbf{95.13\%} & \textbf{91.27\%} & \textbf{93.09\%}             & 220.09      / 73                                                      \\ \hline
\end{tabular}
\end{adjustbox}
\end{table}

In Table~\ref{Tab: AC_results} we can see that the high accuracy levels reached (above 90\%) did not guarantee accurate predictions, as the data imbalance led to biased models. This is evident when analyzing the precision, recall, and F1 score, where biased models had low scores hovering around 50\%, revealing the impact of imbalanced predictions.
During training from scratch, only the VGG16 model appeared unbiased. This observation suggests that smaller models and models with complex novel layers are more susceptible to bias when trained from scratch.

On the other hand, fine-tuned models with pre-trained weights exhibited better resistance to bias, as evidenced by the recall scores. All pre-trained models reported recall rates greater than 78\%, indicating the accurate classification of images into the two labels. Notably, ConvNeXT stood out with an impressive accuracy of 97.60\% and an F1 score of 93.09\%, making it the best-performing model among those we tried.

The reported training times were again consistent with those observed during the training of other tasks. This consistency is attributed to the use of the same number of images and models across all tasks.

\subsection{Curated vs. non-curated data}

As discussed in Section \ref{processing}, the curated MTF data underwent meticulous manual processing whereby a large amount of low-quality, inappropriate, or noisy images were filtered out. To illustrate the benefits of this manual processing, especially for training classification models, in this section we compare the performance of the best-performing DL model (ConvNeXT with pre-training)
on the four tasks when trained on the two MTF data sets (non-curated and curated).
The non-curated data set was automatically split into a training set encompassing 70\% of the total data, and a validation set that includes the remaining 30\%. 
The results of this experiment are presented in Table \ref{Tab: FR_unprocessed}.

\begin{table}[ht]
\centering
\caption{Comparison of ConvNeXT with pre-training on the non-curated MTF data set vs the curated MTF data set}
\begin{adjustbox}{width=0.48\textwidth}
\label{Tab: FR_unprocessed}
\begin{tabular}{@{}llrrrrc@{}}
\hline
Task                & Training data & Accuracy & Precision & Recall  & F1 score & \begin{tabular}[c]{@{}c@{}}Training time\\ (minutes)/\ \\ number \\ of epochs\end{tabular} \\ \hline\hline
\multirow{2}{*}{FR} & Non-curated   & 10.07\%  & 8.85\%    & 8.09\%  & 7.01\%                    & 4089.72  / 94                                                           \\
                    & Curated           & 79.87\%  & 76.29\%   & 73.31\% & 74.07\%           & 201.74    / 76                                                          \\\hline
\multirow{2}{*}{RC} & Non-curated   & 65.38\%  & 59.89\%   & 58.67\% & 59.18\%                & 3809.95    / 87                                                          \\
                    & Curated           & 95.77\%  & 94.41\%   & 94.94\% & 94.65\%                                  & 207.07      / 70                                                        \\\hline
\multirow{2}{*}{GC} & Non-curated   & 69.46\%  & 69.38\%   & 69.32\% & 69.34\%                  & 4094.41   / 93                                                          \\
                    & Curated           & 98.88\%  & 98.86\%   & 98.90\% & 98.88\%                                    & 219.22   / 74                                                            \\\hline
\multirow{2}{*}{AC} & Non-curated   & 62.57\%  & 61.70\%   & 61.57\% & 61.62\%                 & 3911.01    / 89                                                         \\
                    & Curated           & 97.60\%  & 95.13\%   & 91.27\% & 93.09\%                                     & 220.09   / 73\\\hline                                                          
\end{tabular}
\end{adjustbox}
\end{table}

To enable a fair comparison for the FR task, we only considered the 240 identities in the curated MTF data set when training the model on the non-curated version. The ConvNeXT model trained on the non-curated data achieved a mere 10\% accuracy. In contrast, the model trained on the curated data exhibited an impressive 79.87\% accuracy in the 240-label classification. The disparity in performance is evident from the F1 score, where the model trained on the curated data outperformed the model trained on the non-curated data by an order of magnitude. The observed discrepancy in performance was particularly expected in the FR task, given the intended use of both data sets, the key role of data processing in the curated data to discard a large number of crawled images that did not correspond to the correct identity, and the substantial amount of noise in the raw data. These factors explain the large gap in results between the two data sets. From these results, it is clear that only the curated MTF data set should be used for FR tasks.

We limited the comparison on the non-curated data set to the ConvNeXT model, as it consistently outperformed the other models in the curated data experiments across all tasks. This choice was made to provide a strong upper-bound baseline while avoiding the expensive computational costs associated with retraining multiple models on a much larger dataset.

The reported results are based on a single training run per model and task. While this approach provides a consistent baseline for comparison, it does not capture the variability that may arise due to random weight initialization or mini-batch sampling. Researchers may consider running each experiment multiple times with different seeds and reporting averages and standard deviations to better assess the robustness and stability of their model. 
 
For the remaining three tasks (race, gender, and age classification) we observe smaller differences. The model trained on the non-curated data reported accuracy levels ranging from 62\% to 69\%, while the model trained on the curated data reported accuracies higher than 95\%.
Although these three tasks were comparatively easier for the models to learn, and the occurrence of incorrect images crawled from the search engine was low, the impact of manual processing in improving the performance of classification models was evident.

Training runtimes were also proportional to the size of the data. In this case, the non-curated data set resulted in around 20 times higher training costs than the curated data. 

In summary, whereas the curated MTF data set should be used when model performance in data classification is the main goal, the non-curated version would be better suited for data intensive and unsupervised tasks, such as building general or generative AI models. 

\section{Ethical and legal considerations}
The MTF data sets were compiled with strict adherence to ethical and legal requirements, particularly those outlined in the EU GDPR. To ensure a lawful image collection, we exclusively sourced face images of publicly recognized celebrities who had voluntarily made their images publicly available under permissive licenses. Only images with explicit copyright permissions, public domain or Creative Commons licenses that allow modification and reuse were included.

Furthermore, the use of facial images was approved by the BOEL of the SoBigData++ H2020 project. This ethical oversight ensured that the data collection, processing, and labeling procedures were in accordance with established guidelines for privacy, fairness, and transparency. The curated version of the dataset further emphasizes quality and appropriateness by manually excluding images that were synthetic, misleading, or potentially harmful.

\section{Conclusions and future work}
\label{sec: conc}

We have presented two Multi-Task Faces (MTF) data sets, one of them being the curated version of the other, and both containing labeled images of human faces. 
Unlike similar data sets consisting of human faces which are at risk of being removed from the internet for privacy reasons, MTF data were compiled with a view to longevity thanks to compliance with the legal requirements of the GDPR and strict ethical considerations. MTF data exclusively focus on publicly recognized celebrities' images that were either public-domain or published under Creative Commons licenses granting permission for modification, sharing, and commercial use. In this paper, we have elaborated on the data preparation processes, while also presenting comprehensive statistical information and empirical results that aid to better understand the quality and suitable uses of each data set.

We applied the curated MTF data set to four classification tasks using five well-known deep learning models. The ConvNeXT model achieved the best performance in all tasks, with an accuracy of 98.88\% for gender classification, 95.77\% for race classification, 97.60\% for age classification and 79.87\% for face recognition. These results demonstrate that the curated MTF data set offers high-quality labels suitable for training robust AI models. In contrast, the non-curated version showed significantly lower performance, underscoring the importance of careful data curation in building effective and fair machine learning pipelines.

As a future work, we plan to apply and test the MTF data sets to other tasks, such as face anonymization. In particular, the FR task in the curated MTF data set can be used to measure the residual re-identification risk of image anonymization algorithms, whereas the accuracy achieved for the other three tasks on the anonymized images would quantify the degree of utility preserved. In addition, we aim to test the ability of the non-curated MTF data set to train generative models of synthesized human faces.

\section*{Acknowledgment}
The authors appreciate the effort and generosity of the owners of the images we used to make these public domain or release them under the most flexible Creative Commons license.
The owners of the images are acknowledged in a CSV file named ``URL\_of\_original\_images'' and included in the zip file of both MTF data sets that links all the original sources of the images.\\

\bibliographystyle{IEEEtran}
\bibliography{bliography}
\phantomsection 
\begin{IEEEbiography}[{\includegraphics[width=1in,height=1.25in,clip,keepaspectratio]{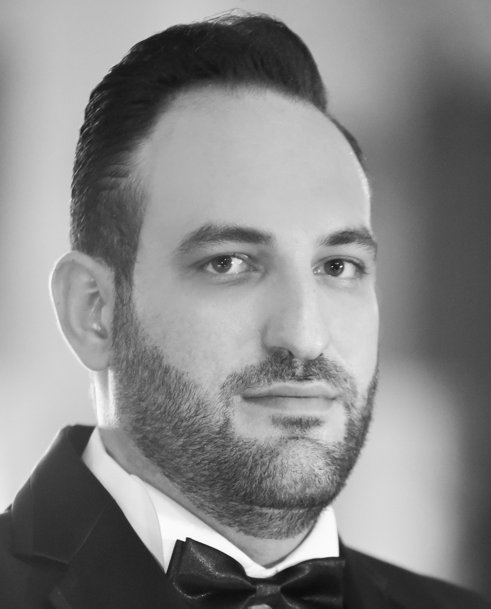}}]{Rami Haffar}
received the MSc and PhD degrees in computer science
from Rovira i Virgili University, Tarragona, Catalonia, in 2019 and 2023, respectively. He is currently a postdoctoral researcher at Rovira i Virgili University. His research interests include explanable AI, machine learning, and image anonimization.
\end{IEEEbiography}

\begin{IEEEbiography}[{\includegraphics[width=1in,height=1.25in,clip,keepaspectratio]{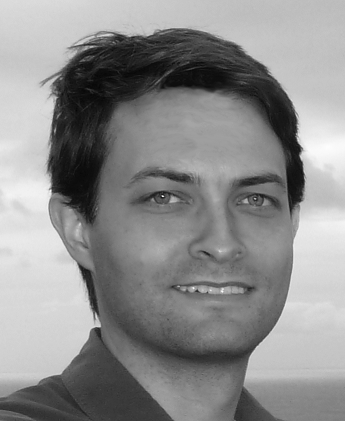}}]{David Sánchez}
(Senior Member, IEEE) received the PhD degree in computer science from the Technical University of Catalonia, Barcelona, in 2008. He is currently a
full professor and an ICREA-Academia researcher with Universitat Rovira i Virgili, Tarragona, Catalonia. He has participated in several
National and European funded research projects
and has authored several papers and conference
contributions. His research interests include data
semantics, ontologies, machine learning, and
data privacy.
\end{IEEEbiography}

\begin{IEEEbiography}[{\includegraphics[width=1in,height=1.25in,clip,keepaspectratio]{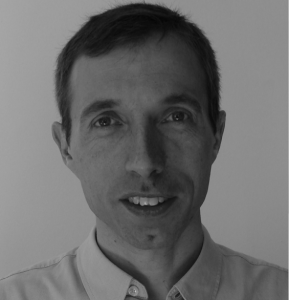}}]{Josep Domingo-Ferrer}
(Fellow, IEEE) received the MSc and PhD degrees in computer science
from the Autonomous University of Barcelona in 1988 and 1991, respectively. He also holds an MSc degree in mathematics. He is currently a distinguished professor of computer science and an ICREA-Academia researcher with Universitat Rovira i Virgili, Tarragona, Catalonia, where he also leads CYBERCAT. His research interests include data privacy, data security, and cryptographic protocols.
\end{IEEEbiography}


\EOD

\end{document}